\documentclass[letterpaper, 10 pt, conference]{ieeeconf}

\IEEEoverridecommandlockouts

\usepackage{multicol}

\usepackage{amsmath}
\usepackage{amssymb}
\usepackage{booktabs}

\usepackage{bm}
\usepackage{booktabs}
\usepackage{multirow}

\usepackage{float}
\usepackage{caption} 
\usepackage{subfigure}

\usepackage{amsfonts}
\usepackage{algorithmic}
\usepackage{graphicx}
\usepackage{textcomp}
\usepackage{todonotes}
\usepackage{csquotes}
\usepackage{xspace}
\usepackage{siunitx}

\usepackage{multicol}
\usepackage{multirow}

\usepackage{lipsum}
\let\labelindent\relax
\usepackage[shortlabels, inline]{enumitem}
\usepackage{wrapfig}

\usepackage{makecell}
\usepackage{hyperref}

\usepackage{cite}

\usepackage{pifont}
\usepackage{subcaption}

\newcommand{\optigrasp}{\text{OptiGrasp}\xspace}
\newcommand{\dexnet}{\text{DexNet3.0}\xspace}

\newcommand{\yi}[1]{}
\newcommand{\markus}[1]{}
\newcommand{\soofiyan}[1]{}

\title{\LARGE \bf \optigrasp: Optimized Grasp Pose Detection Using RGB Images for Warehouse Picking Robots}

\author{Soofiyan Atar$^1$, Yi Li$^1$, Markus Grotz$^1$, Michael Wolf$^2$, Dieter Fox$^1$, Joshua Smith$^1$ 
\thanks{*This work was supported by UW + Amazon Science Hub}
\thanks{$^{1}$SA, YL, MG, DF, and JS are from the University of Washington}
\thanks{$^{2}$Michael Wolf is from Amazon Robotics}
}

\begin{document}

\maketitle
\thispagestyle{empty}
\pagestyle{empty}

\begin{abstract}

In warehouse environments, robots require robust picking capabilities to manage a wide variety of objects. Effective deployment demands minimal hardware, strong generalization to new products, and resilience in diverse settings. Current methods often rely on depth sensors for structural information, which suffer from high costs, complex setups, and technical limitations. Inspired by recent advancements in computer vision, we propose an innovative approach that leverages foundation models to enhance suction grasping using only RGB images. Trained solely on a synthetic dataset, our method generalizes its grasp prediction capabilities to real-world robots and a diverse range of novel objects not included in the training set. Our network achieves an 82.3\% success rate in real-world applications. The project website with code and data will be available at \url{http://optigrasp.github.io}.

\end{abstract}

\section{INTRODUCTION}

Robust picking is a crucial capability for robots, especially in warehouse environments where they must fetch millions of different objects from shelves. Future intelligent robots must acquire strong capabilities for effective deployment in industry and assist human workers in fetching products. These capabilities include low hardware requirements, generalization to novel products, and robustness in different environments. Despite significant progress in this area, achieving methods that meet these requirements while providing robust performance remains challenging.

Recent research often employs depth information as the primary input of perception to enhance grasp prediction accuracy or to reduce the sim-to-real gap. While depth sensors are widely used, they have drawbacks such as high cost, significant latency, multi-device interference, restricted range and resolution, inaccuracy on transparent and highly reflective object surfaces as well as edges, and insufficient accuracy for detecting tiny textures on objects. These limitations pose significant barriers to their widespread adoption in industrial settings. Despite these challenges, relying solely on RGB input for predicting grasp poses remains difficult as the task of grasping itself relies heavily on the 3D structure of the object, which is not easy to retrieve from a single RGB image; things worsen when it is asked to generalize to objects in categories not included in the training set.

\begin{figure}
    \centering
    \includegraphics[width=0.49\textwidth,keepaspectratio=true]{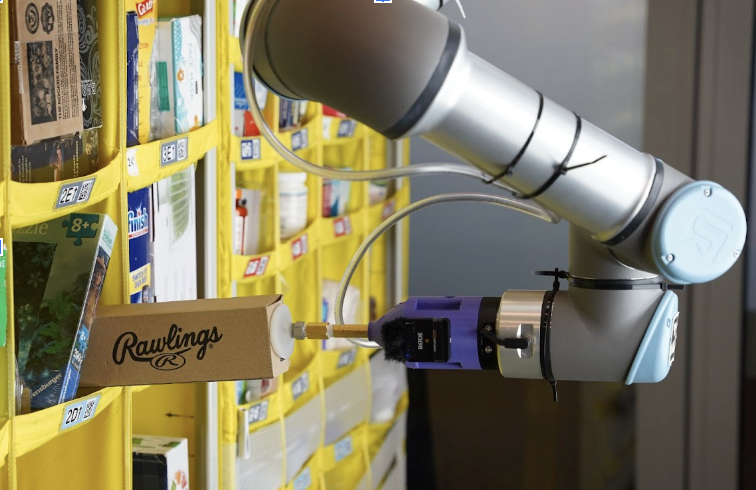}
    \caption{Our robot is picking from a cluttered industrial shelving unit.}
    \label{fig:teaser}
    \vspace{-7mm}
\end{figure}

Inspired by breakthroughs in computer vision, where foundation models demonstrate an understanding of the 3D structure of varies of objects with only RGB images, we introduce a novel approach that leverages the generalization abilities of these models. Our approach aims to improve suction-based robotic grasping by providing detailed affordance maps that guide the robot in selecting the best grasp points and angles. We utilize pre-trained weights from the Depth Anything\cite{yang2024depth} model, a state-of-the-art depth estimation method trained on millions of images, for its capability of understanding 3D structure from RGB images. Following the DINOv2\cite{oquab2023dinov2} backbone and Dense Prediction Transformer \cite{ranftl2021vision} decoder from Depth Anything~\cite{yang2024depth}, we designed the afford grasp head to predict two crucial affordances for grasping: 1. Translation: the affordance map for the best grasp pose, and 2. Rotation: the yaw and pitch angles at which the gripper should approach each grasp point. Although trained exclusively in simulation without any real-world fine-tuning, our network achieves an overall success rate of \textbf{82.3\%} when deployed on the real robot.

Our contributions are summarized as follows:
\begin{itemize}
\item We demonstrate that leveraging pre-trained weights from depth estimation models allows our approach to generalize grasp pose predictions from synthetic training data to unseen real-world objects without any fine-tuning.
\item We propose a simple yet effective network structure to predict grasp poses using a single RGB image, eliminating the need for expensive and complex depth sensors, and introduce an affordance grasp score to efficiently measure the possibility of grasping on each pixel in the image.
\item We generate a large synthetic dataset within a shelf environment containing over 400,000 image data containing 350+ unique objects, which are further domain randomized, featuring high-quality textures on objects. 
\item We conduct extensive real-world evaluations with a diverse set of objects, showcasing our method's superior performance and ability to generalize, achieving an \textbf{82.3\%} success rate in a cluttered warehouse scene.
\end{itemize}

\section{RELATED WORK}

\label{sec:related_work}

\begin{table}[t]
\centering
\caption{Overview of related work. Methods based on either RGB or depth images that output a grasp with translation $\mathbf{t}$ or rotation $\mathbf{R}$.}
\label{tab:related_work}
\begin{tabular}{l|cc|cc}
\hline\rule{0pt}{2ex}
\multirow{2}{*}{Method} & \multicolumn{2}{c|}{Input (visual)} & \multicolumn{2}{c}{Output (grasp)} \\
\cline{2-5}\rule{0pt}{2.2ex}
                        & Modality  & RGB only   & $\mathbf{t}$  & $\mathbf{R}$ \\
\hline\rule{0pt}{2ex}
\hspace{-1mm}SuctionNet \cite{cao2021suctionnet}  & RGB-D   & No   & Yes  & No \\
DexNet 3.0 \cite{mahler2018dex}      & Depth   & No   & Yes  & No \\
Zeng \cite{zeng2022robotic}          & RGB-D   & No   & Yes  & No \\
DYNAMO-GRASP \cite{yang2023dynamograsp} & Depth   & No   & Yes  & No \\
SimSuction \cite{li2024simsuction}   & RGB-D   & No   & Yes  & Yes \\
Ours                                 & RGB     & Yes  & Yes  & Yes \\
\hline
\end{tabular}
\vspace{-5mm}
\end{table}

Suction-based robot manipulators have become increasingly popular in practical applications. For instance, suction grasping techniques are widely-used in manufacturing \cite{zhang2022robotic, yang2021cooperative, olesen2020collaborative}, warehousing \cite{hasegawa2019three, schwarz2017nimbro}, underwater manipulation \cite{stuart2015suction, kumamoto2021underwater}, and food and fruit handling \cite{chua2003robotic, morales2014soft, gilday2020suction, blanes2011technologies}, among other fields. Another significant area where suction grasping is applied involves the exploration of end-effector modalities \cite{jeong2020integration, huh2021multi, mazzolai2019octopus, nakahara2020contact}. \cite{jeong2020integration} introduces a hand exoskeleton equipped with self-sealing suction cup modules to facilitate various grasping tasks. \cite{huh2021multi} discusses a multi-chambered suction cup that supports functions ranging from gentle haptic exploration to detecting seal breaks during strong grips. \cite{mazzolai2019octopus} describes a conical soft robotic arm with suction cups designed to retrieve objects from confined spaces, grasp complex shapes, and operate in diverse environments.
In the following, we will distinguish between analytical methods and learning-based approaches.

\paragraph{Analytic Models}
In the domain of traditional suction cup grippers, effective analysis of grasp quality requires modeling various properties of the cups. Since these suction cups are typically made from elastic materials like rubber or silicone, researchers often use spring-mass systems to represent their deformations~\cite{mahler2018dex, cao2021suctionnet, provot1995deformation}. Once a suction gripper secures a firm grasp on an object, the suction cup is usually modeled as a rigid body. The analysis then focuses on evaluating the forces exerted on the object, including those along the surface normal, friction-induced tangential forces, and suction-generated pulling forces~\cite{kolluru1998modeling}. Mahler et al.\cite{mahler2018dex} introduced a combined model in \dexnet that integrates torsional friction and contact moment within a compliant model of the contact ring between the cup and the object. This combined model has proven effective and is utilized in subsequent works \cite{cao2021suctionnet, mahler2019learning}.  Meanwhile, the Centroid method, a straightforward approach involving suctioning on the object’s centroid, has proven effective in similar tasks at the Amazon Robotics Challenge~\cite{yu2016summary, hernandez2016team}.

\paragraph{Learning Suction Grasps}
Machine learning research in robotics has been actively investigating the selection of optimal grasp points to improve suction grasping for complex manipulation tasks~\cite{jiang2022learning, mousavian20196}. These tasks include picking novel objects, sorting objects, and picking from containers. Existing approaches generate training data either through human expertise~\cite{zeng2022robotic} or simulations~\cite{mahler2018dex, cao2021suctionnet, jiang2022learning, shao2019suction}. For instance, \dexnet~\cite{mahler2018dex} synthesizes training data and proposes suction grasp points to form an effective suction seal and ensure wrench resistance. 
Indeed, a key challenge for learning-based methods is getting high-quality training data. 
Recently, a trend has been to employ large-scale simulation systems, such as DYNAMO-GRASP \cite{yang2023dynamograsp} or SIM-SUCTION \cite{li2024simsuction}, for data generation.
Several other studies focus on clustered scenarios by developing models that take RGB-D input and predict grasp points~\cite{cao2021suctionnet, shao2019suction, zeng2022robotic}. Jiang et al.~\cite{jiang2022learning} proposed a method that simultaneously considers grasping quality and robot reachability for bin-picking tasks. Other aspects include modeling the uncertainty \cite{cao2024uncertainty} or grasping moving objects while also avoiding dynamics obstacles \cite{li2024dbpf}.
Despite these studies primarily focusing on analyzing surface properties or robot configuration, they often require in-depth information or overlook the grasp angle, which might affect the success of the task. Addressing this particular aspect is the main focus of our work. Recent works leverage visual pretraining to improve robotic manipulation, enhancing sample efficiency and performance in tasks like grasping and object manipulation. Techniques include using affordance maps, masked autoencoders, and video-language alignment to create robust visual representations that facilitate faster learning and better transferability across different tasks \cite{nair2022r3muniversalvisualrepresentation, radosavovic2022realworldrobotlearningmasked, xiao2022maskedvisualpretrainingmotor, yenchen2021learninglearningactvisual}.

\begin{figure*}[t]
    \centering
    \includegraphics[width=0.62\textwidth]{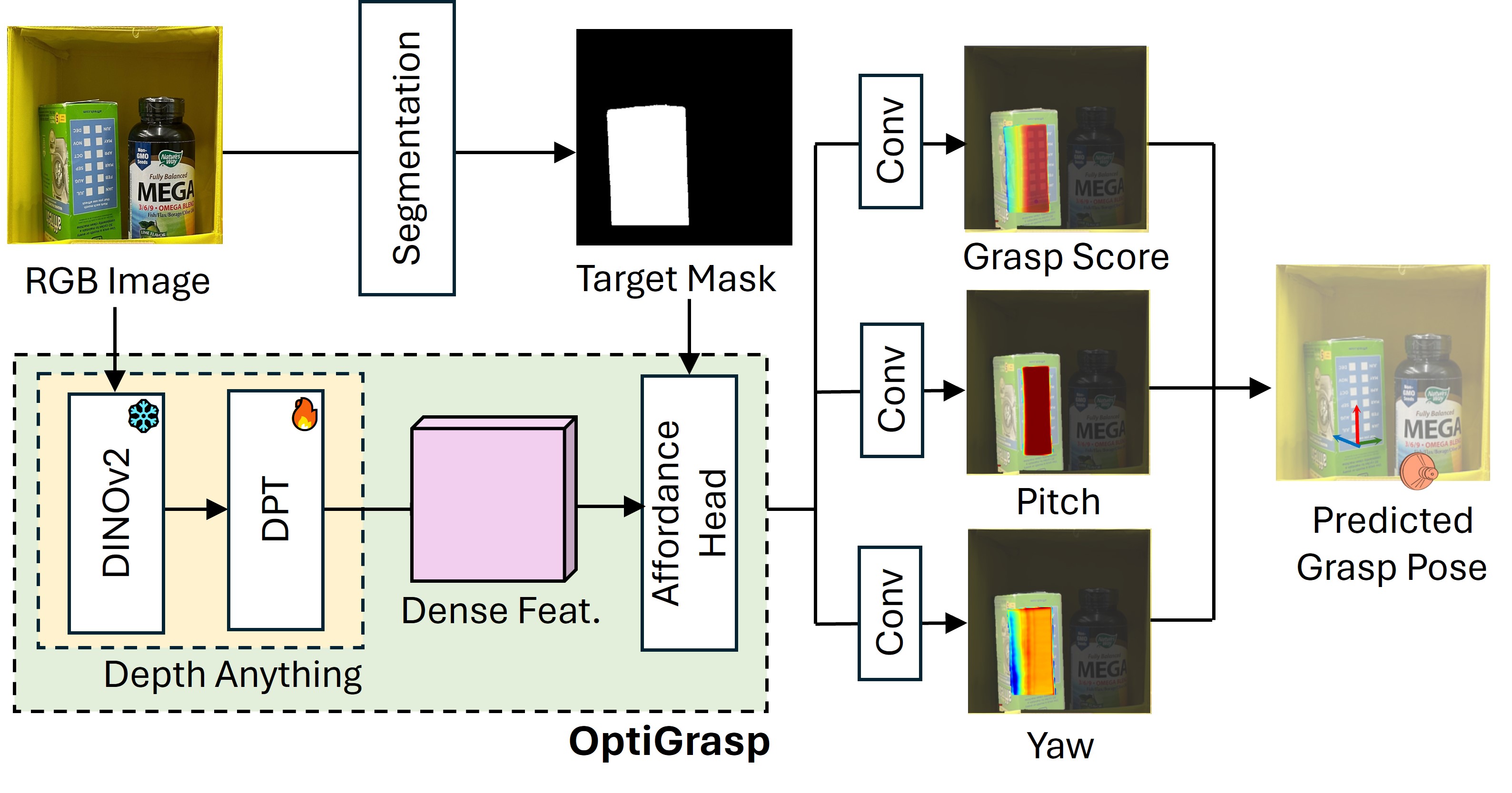}
    \caption{\textbf{The system architecture.} The network takes an RGB image and the mask of the target object as inputs and predicts three dense prediction maps, each of the same size as the input image. These maps predict the affordance grasp score, pitch angle, and yaw angle at each pixel, as described in \autoref{sec:grasp_score}. The higher the value, the redder it is visualized. This prediction is further processed to determine the optimal grasp pose for the suction gripper to pick the object. For the best grasp point, the highest value from the grasp score affordance map is selected, and the corresponding pixel from the pitch affordance map and yaw affordance map is used to compute the final grasp pose. The DINOv2~\cite{oquab2023dinov2} backbone from Depth Anything~\cite{yang2024depth} retains its frozen weights, while the Dense Prediction Transformer (DPT)~\cite{ranftl2021vision} is refined during training.}
    \label{fig:systemarchitecture}
\vspace{-5mm}
\end{figure*}

\section{Problem definition}
\label{sec:problem_definition}
Our objective is to identify optimal grasp points on a target object situated within a container filled with multiple items using only a single-view RGB image. These identified grasp points should allow a robot to successfully establish a suction grasp by selecting objects with favorable geometry and an optimal corresponding approach angle for the gripper. Consistent previous suction grasp point detection studies~\cite{mahler2018dex,cao2021suctionnet,mahler2019learning}, a grasp point is defined by a 6D pose target point $[\mathbf{p}, \mathbf{v}]$, where, $\mathbf{p} \in \mathbb{R}^3$ represents the center of the contact ring between the suction cup and the object, while vector \(\mathbf{v} \in \mathbb{S}^2\), representing the gripper's approach direction, includes pitch \(\beta\) and yaw \(\gamma\). The roll angle is flexible, owing to the symmetry of the suction cup. 

The location can be obtained by projecting the image location with the camera intrinsics $\mathbf{K}$ and extrinsic $(\mathbf{R}, \mathbf{t})$. The depth for the reprojection does not need to be precise since the end effector moves to a pre-grasp pose and then follow the 6D pose waypoints until it either grasps the object or exits the bounds. Hence, depth can be inferred either from the model or by using the location of the bin.
However, it is crucial to accurately determine the location and orientation of the grasp since objects are densely packed on the shelf.

Finally, we make the following assumptions when developing our method \optigrasp:
\begin{itemize}[nosep]
    \item The location of the shelf and its bins are known.
    \item The target object can be identified and segmented by the perception system.
\end{itemize}
The first assumption can be relaxed by scanning the bin's data matrix to locate the object on the shelf.
\vspace{-3mm}
\section{METHOD}
\label{sec:method}

This section describes \optigrasp, a learning-based pipeline developed to create a grasp point detection model. This configuration processes only a single view RGB image of the scene configuration, generating three affordance maps: grasp location, pitch \( \beta \), and yaw \( \gamma \) on the target object or all the objects within the scene. The primary map estimates the likelihood of successful suction grasp, while the additional maps predict the best roll and yaw angles for optimal grasping points identified on the primary map. These maps collectively predict the pixel-wise success probability of object pickup, as illustrated in \autoref{fig:systemarchitecture}. Note that the model was trained exclusively on synthetic images without any real-world fine-tuning, and zero-shot transfer was demonstrated in real-world experiments.

\vspace{-1mm}
\subsection{Network Structure}
The \optigrasp as shown in \autoref{fig:systemarchitecture} integrates a pre-trained DINOv2~\cite{oquab2023dinov2} from the Depth Anything~\cite{yang2024depth} model using a ViT-base architecture. This pre-trained network processes input single-view RGB images to generate a dense feature map. The output from DINOv2 is subsequently passed to a DPT model. The output from the DPT~\cite{ranftl2021vision} model is then combined with the segmentation mask obtained from STOW~\cite{li2023stow}, provides the segmentation masks, and tracks unseen object instances in discrete frames. It is then passed to the Affordance Grasp Head. This module produces three affordance maps corresponding to the grasp point, pitch \( \beta \), and yaw \( \gamma \), facilitating the interpretation of scene affordances for determining the 6D pose.

The system operates on single-view synthetic RGB images. Segmentation masks are incorporated into the Affordance Grasp Head for training. The system calculates loss across all segmented pixels for each affordance map as referred in \autoref{eqn:loss_eqn}.

We adopt sim-to-real transfer to predict real-world visual affordances without direct fine-tuning on real-world images as they are expensive to collect. The approach is restricted to single-view RGB images, utilizing depth images solely for label formation. The resultant 6D pose comprises a grasp location vector ($\mathbf{p}$) along with \( \beta \) and \( \gamma \) angles derived from the affordance maps. \( \beta \) and \( \gamma \) are chosen based on the specific point that corresponds to the highest grasp score in the evaluation. Both angles are constrained within (-30, 30) degrees, owing to bin constraints and the challenges posed by dynamic scenarios. The system tolerates up to ±15 degrees for highly tilted objects through suction deformation beyond the constrained range. Labels are generated every 5 degrees within (-30, 30) degrees as increasing the resolution leads to longer data generation time. During the grasp point selection process, we select the highest score on the grasp location affordance map and extract corresponding points from the pitch and yaw maps to determine the optimal grasp configuration.

\vspace{-2mm}
\subsection{Simulation Environment and Data Generation}
To avoid costly real robot data collection, we developed a simulated environment for data generation. The pipeline has a similar setup with a vertical shelf arrangement as from DYNAMO-GRASP~\cite{yang2023dynamograsp} setup, incorporating 350+ diverse object sets from Google's scanned objects~\cite{downs2022google}. In our simulation setup, the number of objects placed in the bin for each scene configuration is determined randomly based on the bin's volume, subject to space limitations. Each object is then positioned within the bin using random translations within the bin bounds and random rotations. Following placement, we collect depth data, segmentation information, and single-view RGB data from the scene. After collecting the data, the scene is reset for the next configuration. This process ensures efficient data generation for training purposes. In the \optigrasp framework, the RGB image is the primary input, while the segmentation mask is only used to outline the target object.  Additionally, noiseless depth images are employed to generate specific labels for the object in question. Domain randomization is used to vary friction coefficients, object sizes, and weights, enhancing model robustness and generalization to real-world scenarios. In the simulation setup, the initial configuration includes 30 objects, with convex decomposition applied to both the pod and the objects to obtain fine-grained collision models. This setup allows for spawning only 30 objects at once across 75 different environments in parallel. After collecting data from every 10 scene configurations, a new set of 30 objects is introduced, maintaining variability in object shapes. Object sizes and weights vary randomly with each new spawn to ensure diversity in the simulation parameters.
\begin{figure*}[t]
    \centering
    \includegraphics[width=0.75\textwidth]{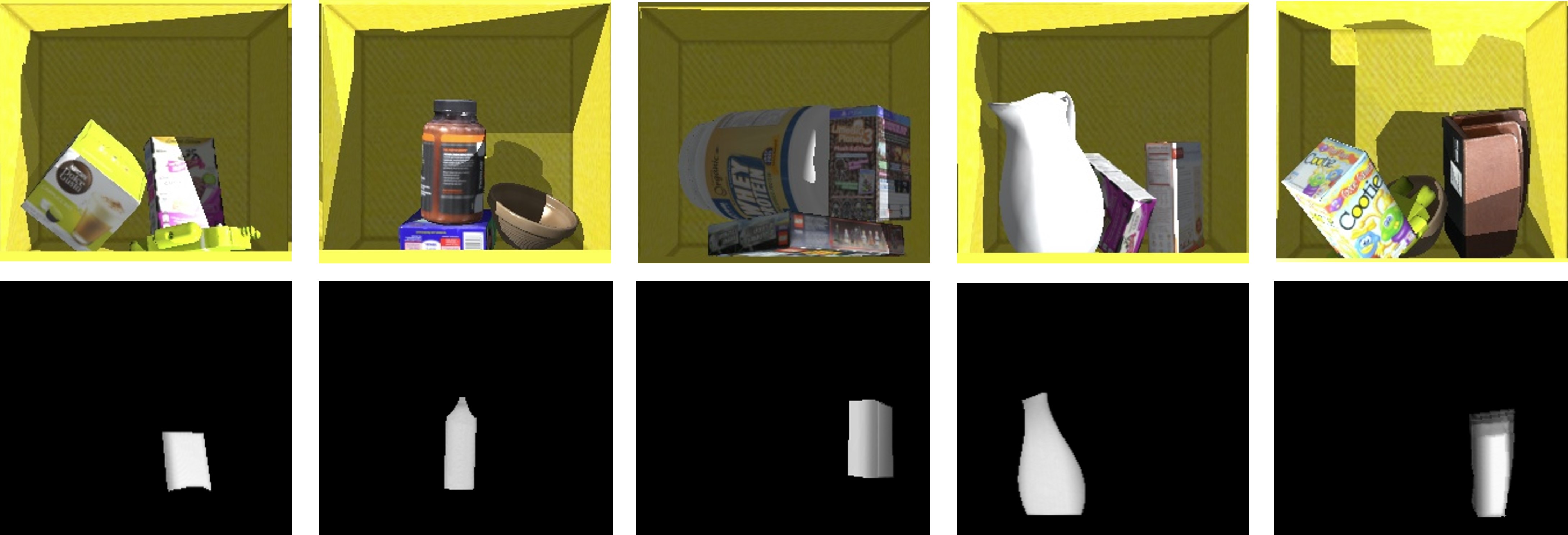}
    \caption{\textbf{Illustration of the synthetic data we generated.} The first row shows RGB images, while the second-row lists per-pixel affordance scores computed with the affordance grasp score in \autoref{eq:deformation_cost}}
    \label{fig:synthetic_data}
    \vspace{-5mm}
\end{figure*}

\subsection{Data Labelling}
\label{ref:labelling_data}
The labeling process involves evaluating the \( F \) for a wide set of pitch $\beta $ and yaw $ \gamma $ angle combinations for each object. The point cloud is rotated for each set to align with the suction cup's approach angle. This alignment ensures that the calculated scores reflect the actual approach of the suction cup.

The affordance grasp score \( F (\beta, \gamma) \) described in \autoref{sec:grasp_score}  is computed for each object pixel within the rotated point cloud. The dataset comprises approximately 400,000 instances, and for each instance, \( F \) is determined for every combination of \( \beta \) and \( \gamma \). The best \( F (\beta, \gamma) \) for each object is identified by evaluating all angle configurations and updating the \( F \) if a new configuration yields a better \( F (\beta, \gamma) \). Consequently, three affordance labels for each object configuration are generated, capturing the best \( \beta \), \( \gamma \), and corresponding \( F \).

To generate the labels efficiently, we utilized eight NVIDIA A10G Tensor Core GPUs with multi-GPU parallelization, enabling the collection of labels in a highly parallelized manner. This setup simplified the labeling process, providing a dataset for training and validation.

\subsection{Affordance Grasp Score}
\label{sec:grasp_score}
The affordance grasp score \( F \) for each resolution of pitch $\beta$ and yaw $\gamma$ angles is defined as:

\vspace{-5mm}
\begin{equation}
F(\beta, \gamma) = k_1 S_a - k_2 C_d - k_3 V_d + k_4 S_n + k_5 S_c
\label{eq:deformation_cost}
\end{equation}

where \( S_a \) is the normalized anomaly score, \( C_d \) is the depth consistency cost, \( V_d \) is the depth variability cost, \( S_n \) is the normal consistency score, and \( S_i \) is the angle inclination score. The weighting factors \( k_1, k_2, k_3, k_4, k_5 \) are positive values balancing the importance of each component. The scores are calculated as shown in \autoref{tab:suction_def_score}.

\begin{table}[h!]
\centering
\begin{tabular}{ll}
\toprule
\textbf{Score} & \textbf{Equation} \\ \midrule
Normalized Anomaly Score (\( S_a \)) & \( S_a = \left(1 - \frac{\sum_{i=1}^{N} (D_{max} - D_i)}{A_{\text{max}}}\right) \) \\
Depth Consistency Cost (\( C_d \)) & \( C_d = \Delta \theta \sum_{i=1}^{N} \left| D_i - D_{i+\Delta \theta} \right| \) \\
Depth Variability Cost (\( V_d \)) & \( V_d = \sigma_D \) \\
Normal Consistency Score (\( S_n \)) & \( S_n = \frac{1}{N} \sum_{i=1}^{N} \left( \frac{\theta_{\text{thresh}} - \theta_i}{\theta_{\text{thresh}}} \right) \) \\
Angle Inclination Score (\( S_c \)) & \( S_c = \left( \frac{\theta_{\text{max}} - \theta_i}{\theta_{\text{max}}} \right) \) \\
\bottomrule
\end{tabular}
\caption{\textbf{Definitions for different scores used in the affordance grasp score function.} \( N \) is the number of depth measurements along the perimeter suction projection, \( D_i \) is the depth at point \( i \) along the perimeter of the suction projection, \( D_{max} \) is the maximum depth, \(A_{\text{max}}\) is the maximum Anomaly score, \( \Delta \theta \) is the angle resolution, \( \sigma_D \) is the standard deviation of the depth values, \( \theta_i \) is the angle between the normal vector at point \( i \) and the reference normal vector, and \( \theta_{\text{thresh}} \) and \( \theta_{\text{max}} \) is the threshold and maximum allowed inclination angles, respectively. These components collectively ensure robust and reliable suction grasps.}
\label{tab:suction_def_score}
\vspace{-6mm}
\end{table}

\subsection{Training}
We trained the method for 70 epochs using 8 NVIDIA Tesla V100 GPUs with a total of \SI{128}{\giga\byte} GPU memory. We selected the Adam optimizer and added a scheduler to adjust the learning rate. \optigrasp was trained exclusively on Synthetic single-view RGB Images, and we adopted sim-to-real transfer in a zero-shot style and evaluated it in the real world.

\vspace{-5mm}
\begin{equation}
L_{\delta}(a) = \sum_{i \in \mathcal{A}} \sum_{j \in \text{Mask}} \begin{cases} 
\frac{1}{2} (y_{ij} - \hat{y}_{ij})^2 & \text{if } |y_{ij} - \hat{y}_{ij}| < \delta \\
\delta (|y_{ij} - \hat{y}_{ij}| - \frac{1}{2} \delta) & \text{otherwise} 
\end{cases}
\label{eqn:loss_eqn}
\end{equation}
We use Huber loss \(L_{\delta}(a) \) as shown in \autoref{eqn:loss_eqn} to calculate the loss for all three affordance maps, where \(y\) represents the labels which are produced as mentioned in \autoref{ref:labelling_data} and \(\hat{y}\) represents the affordance maps predictions. The loss is calculated by summing all three affordance maps for grasp location, pitch, and yaw. \(\delta\) is a threshold parameter that is set to 1.0 throughout this paper. $\mathcal{A}$ represents the set of all affordance maps and $\text{Mask}$ represents pixels within the segmentation mask.

\vspace{-2mm}
\section{EXPERIMENTS}
\subsection{Real Robot Setup}

\begin{figure}[bt]
    \centering
    \includegraphics[width=0.65\linewidth,keepaspectratio=true]{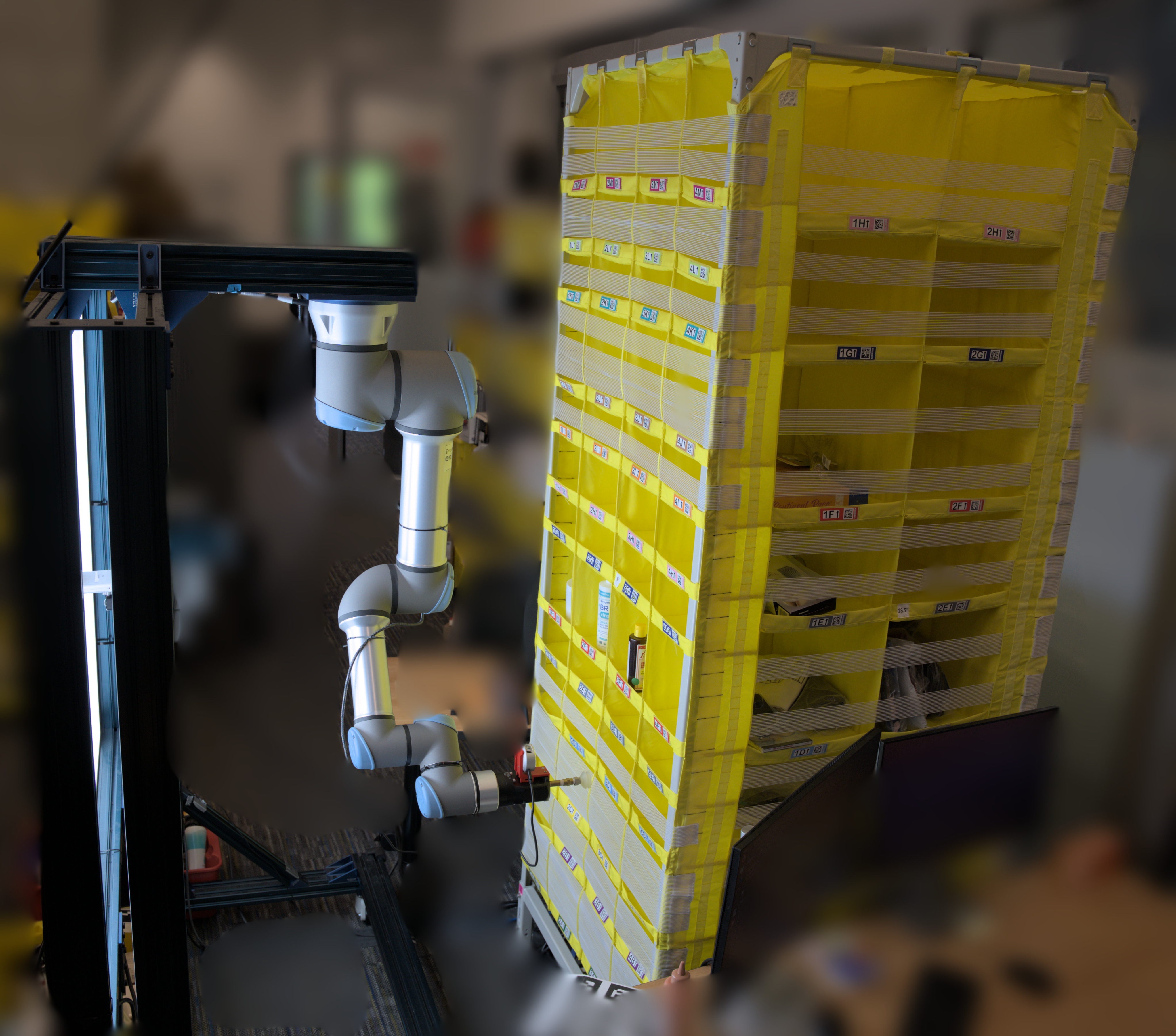}
    \caption{Robotic work cell.}
    \label{fig:experimentsetup}
    \vspace{-5mm}
\end{figure}

The objective of our experiment is to investigate robotic suction grasping for industrial warehouse shelves, as detailed in \cite{Grotz2023rss}. 
\autoref{fig:experimentsetup} depicts the robot setup and the industrial shelving unit, where a huge variety of objects can be stored.
Throughout the evaluation, a Universal Robots UR16e robot was equipped with a custom industrial vacuum suction gripper. The vacuum gripper houses a Schmalz SCTSi-EIP 4  vacuum ejector, which has a maximum flow rate of \SI[per-mode=repeated-symbol]{65.5}{\liter\per\minute}. RRT Connect\cite{kuffner2000rrt} along with OMPL\cite{sucan2012open} was used for motion planning of the robotic arm. 

In our experiment, we benchmark three methods under the same setup: 1. Our method \optigrasp; 2. \dexnet~\cite{mahler2018dex}, and 
3. The centroid method. 
\dexnet is a state-of-the-art suction-picking technique, serving as a strong baseline. Meanwhile, the Centroid method, a straightforward approach involving suctioning on the centroid of the object mask, has proven effective in similar tasks at the Amazon Robotics Challenge~\cite{hernandez2017team, yu2016summary}. SimSuction~\cite{li2024simsuction} is trained on top-down scenarios and uses Isaac Sim for data collection related to object dynamics. Since our method does not rely on this approach, a direct comparison would require significant architectural changes, making it an unfair comparison.

To simulate a realistic warehouse environment, we used a diverse range of objects with varying shapes and properties. The object sets were categorized based on difficulty into easy, medium, and challenging levels to assess our method's efficacy.

\begin{figure*}[t]
    \centering
    \includegraphics[height=0.3\textwidth]{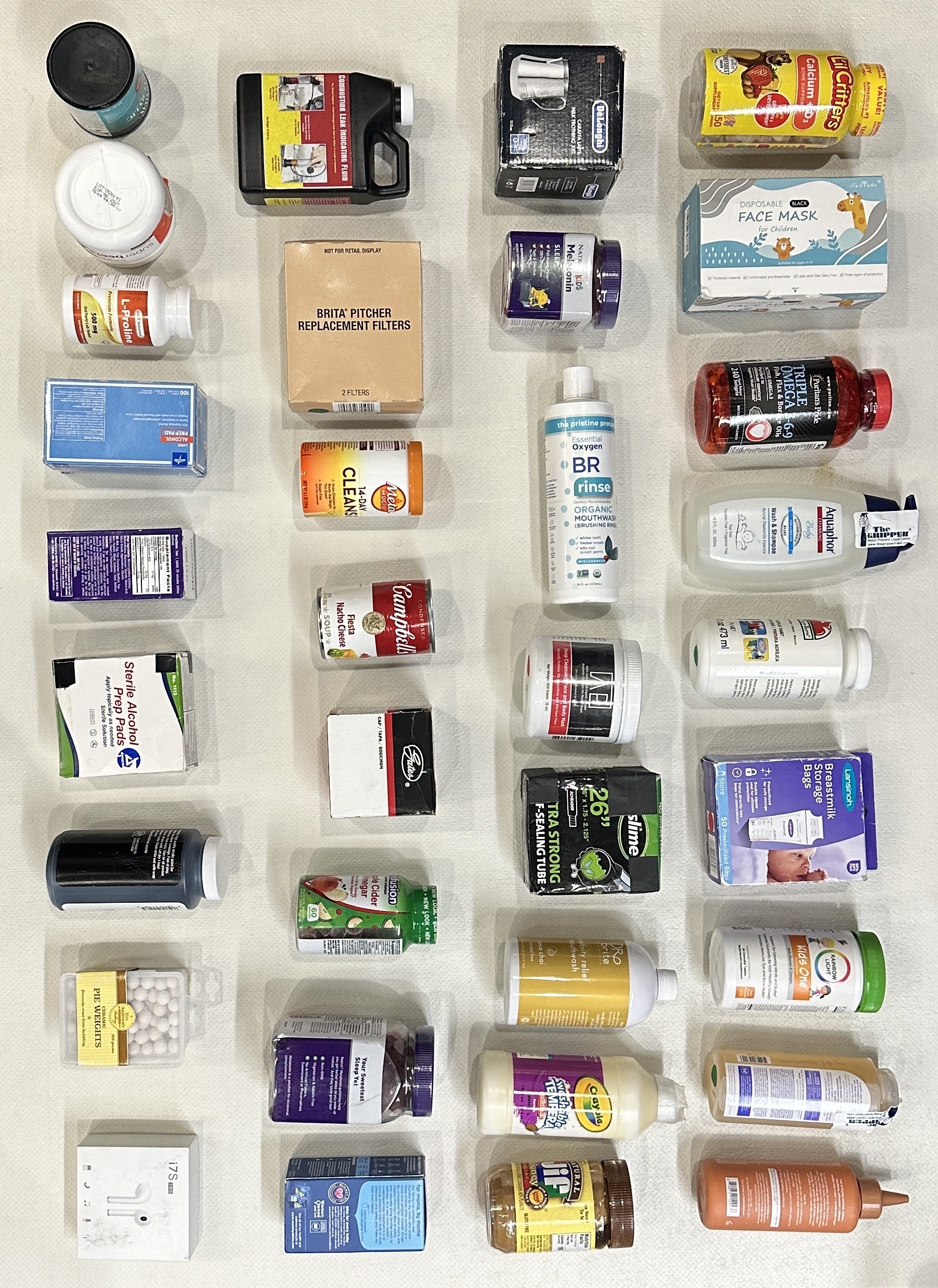}
    \includegraphics[height=0.3\textwidth]{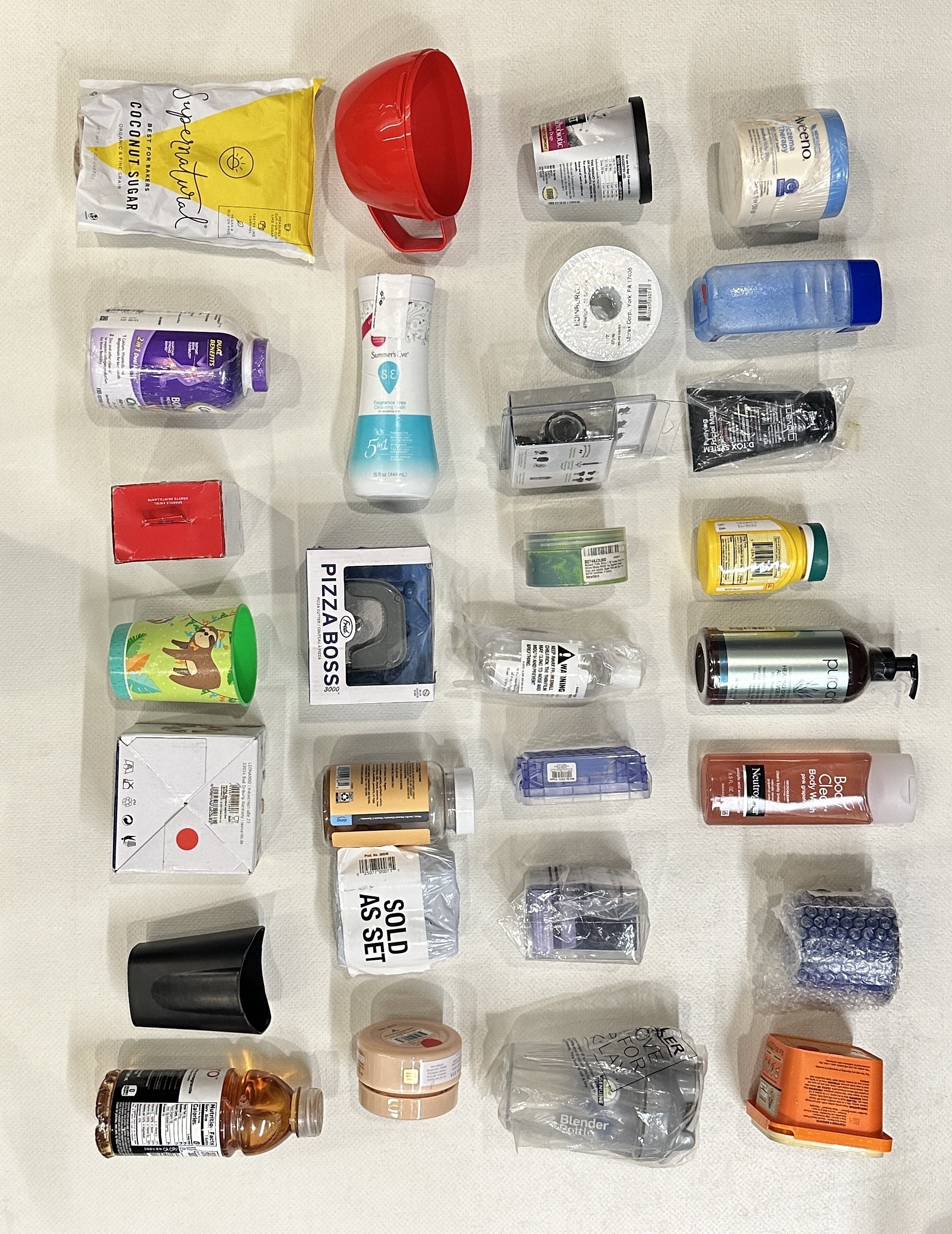}
    \includegraphics[height=0.3\textwidth]{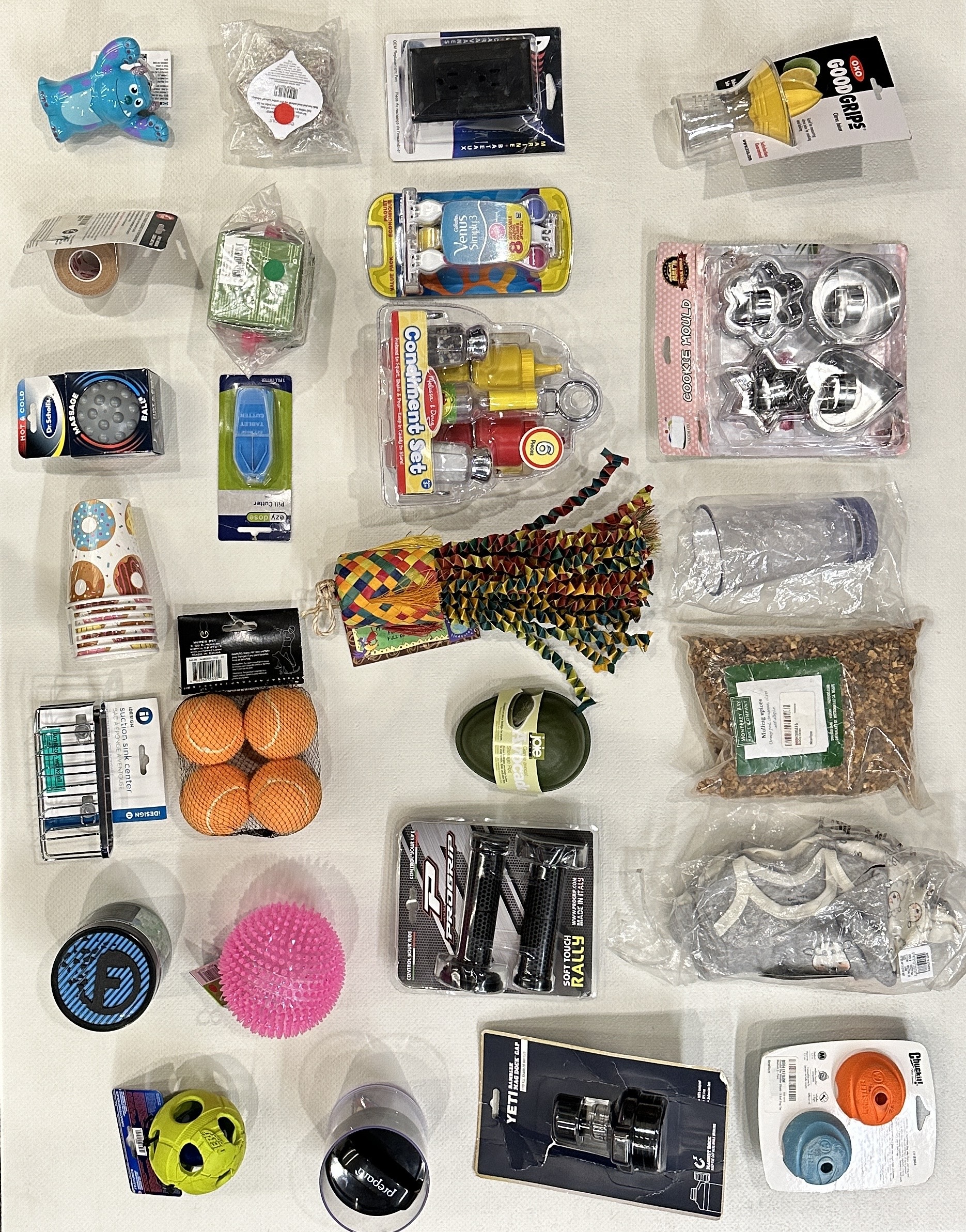}
    \caption{Our three object sets range from easy, medium to Hard (left to right)}
    \label{fig:object_set}
    \vspace{-1mm}
\end{figure*}

\autoref{fig:object_set} displays the categorized object sets:

\textbf{Easy Object Set:}Consists of bottles and boxes, straightforward to grasp but may pose orientation challenges

\textbf{Medium Object Set:}Contains bottles and boxes with geometric irregularities and transparent sections, complicating grasping.

\textbf{Hard Object Set:}This includes objects with minimal graspable areas or deformable materials that are unsuitable for suction-based grasping due to hardware limitations.

We measure the grasp success rate, which is defined as the number of successful picks divided by the number of pick attempts.

\subsection{Results}

\begin{table}[t]
\centering
\caption{Success rates of grasping methods across object sets of different difficulty. The evaluation involved 215 grasps on 170 unique objects.}
\label{tab:grasping_methods_real}
\setlength{\tabcolsep}{3pt}
\begin{tabular}{l|cccccc}
\hline\rule{0pt}{2ex}
Method & Input & Easy & Medium & Hard & \makecell{Objects \\ Grasped} & \makecell{Grasp \\ Accuracy} \\
\hline\rule{0pt}{2.5ex}
\hspace{-2mm}\makecell{OptiGrasp} & RGB & \textbf{90.9\%} & \textbf{82.7\%} & \textbf{73.3\%} & \textbf{176/215} & \textbf{81.9\%}\\
\hspace{-1mm}\makecell{OptiGrasp \\ w/o angle} & RGB & 83.1\% & 65.4\% & 54.7\% & 145/215 & 67.4\%\\
\hspace{-1mm}\makecell{Centroid} & RGB & 77.9\% & 61.5\% & 37.2\% & 123/215 & 57.2\%\\
\hspace{-1mm}\makecell{DexNet 3.0} & Depth & 68.8\% & 50.0\% & 20.9\% & 87/215 & 40.5\% \\
\hline
\end{tabular}
\vspace{-1mm}
\end{table}
\autoref{tab:grasping_methods_real} summarizes the results on the real robot. 
To make the results reproducible, we restored the state of the perception system after each trial and carefully placed the objects back in the same location in the bin after each attempt. A pick attempt is counted as successful if the seal of the suction cup is closed and the object is lifted from the bin.  In total, \optigrasp made 215 pick attempts for three different object sets with 176 successful picks and an 82.3\% success rate. For easy objects which are mainly boxes and bottles, it achieves a success rate of over 90\%. It's notable that removing the pitch and yaw angle has a significant negative impact on \optigrasp's performance.

For the evaluation of synthetic data, grasp success is determined by comparing results with generated labels; if they fall within the threshold, it is classified as successful. The results are shown in \autoref{tab:model_weight_strategy}. It also shows that our method outperforms all other methods on synthetic dataset evaluation.

\begin{table}[t]
\centering
\caption{Accuracy comparison of grasping methods using RGB and depth data across 230 real-world objects. Our method (OptiGrasp) achieves better accuracy.}
\label{tab:dexnet_variant}
\begin{tabular}{l|c|c}
\hline\rule{0pt}{2ex}
Model & Input & Grasp Accuracy \\
\hline\rule{0pt}{2.5ex}
\hspace{-1mm}OptiGrasp (Ours) & RGB & \textbf{78.0\%} \\
DexNet 3.0 & \makecell{Depth from \\ Depth Anything \cite{yang2024depth}} & 54.3\% \\
DexNet 3.0 & \makecell{Depth from \\ depth camera} & 50.4\% \\
DYNAMO GRASP & \makecell{Depth from \\ depth camera} & 48.5\% \\
\hline
\end{tabular}
\vspace{-5mm}
\end{table}

We provided DexNet with depth data from the Depth Anything model, depth from a depth camera, and OptiGrasp with RGB input, evaluating 230 objects via human expert assessment. As shown in Table \ref{tab:dexnet_variant}, DexNet's performance improved with Depth Anything data, but the accuracy gain over camera depth was minimal. Thus, depth data alone is insufficient for accurate grasp point prediction. DYNAMO GRASP \cite{yang2023dynamograsp}, focused on object dynamics, performed worse and was not evaluated with Depth Anything. Our method leverages a backbone trained on millions of real-world images, facilitating the transfer of grasping skills from synthetic datasets to real-world tasks.

\begin{table*}[tb]
    \centering
    \small 
    \setlength{\tabcolsep}{4pt} 
    \begin{tabular}{l|p{2cm}|l|p{2cm}|c|c}
    \hline\rule{0pt}{2ex}
    Backbone & Fine-Tuned & DPT & Fine-Tuned & Synthetic & Real\\
    & & & & success rate & success rate \\
    \hline
\rule{0pt}{2ex}Depth-Anything\cite{yang2024depth} & No & Depth-Anything\cite{yang2024depth} & Yes & \textbf{79.6} & \textbf{88.9} \\
    Depth-Anything\cite{yang2024depth} & No & Depth-Anything\cite{yang2024depth} & No & 78.4 & 82.3\\
    DinoV2\cite{oquab2023dinov2} & Yes & DPT\cite{ranftl2021visiontransformersdenseprediction} & No & 71.4 & 42.8\\
    DinoV2\cite{oquab2023dinov2} & No & DPT\cite{ranftl2021visiontransformersdenseprediction} & No & 71.0 & 25.0\\
    \hline
    \end{tabular}
    \vspace{0.5\baselineskip}
    \caption{Comparison of accuracy across different backbone and DPT configurations, where "Real success rate" denotes the accuracy achieved during real robot experiments on real-world data, and "Synthetic success rate" represents the accuracy on synthetic data, calculated by comparing model predictions with the true labels (human expert evaluation). The backbone corresponds to the specific DinoV2 variant used in our study where the DinoV2 structure is similar to the original DinoV2\cite{oquab2023dinov2} and From Depth Anything\cite{yang2024depth}, while DPT refers to the variant utilized for ablation analysis.}
    \label{tab:model_weight_strategy}
\vspace{-5mm}
\end{table*}

In Table \ref{tab:model_weight_strategy}, the OptiGrasp architecture outperforms other models in sim-to-real transfer, as evidenced by its superior accuracy in both synthetic and real-world experiments. Conversely, the model trained entirely from scratch exhibits significantly lower performance, indicating the importance of a pre-trained backbone from Depth Anything. Thus, training from sim to real is evident in Opti Grasp, which shows that using a pre-trained backbone helped us increase our performance.

\subsection{Failure Cases and Future Work}

\begin{figure}[t]
    \centering
    \subfigure[]{\centering\includegraphics[height=.12\textwidth]{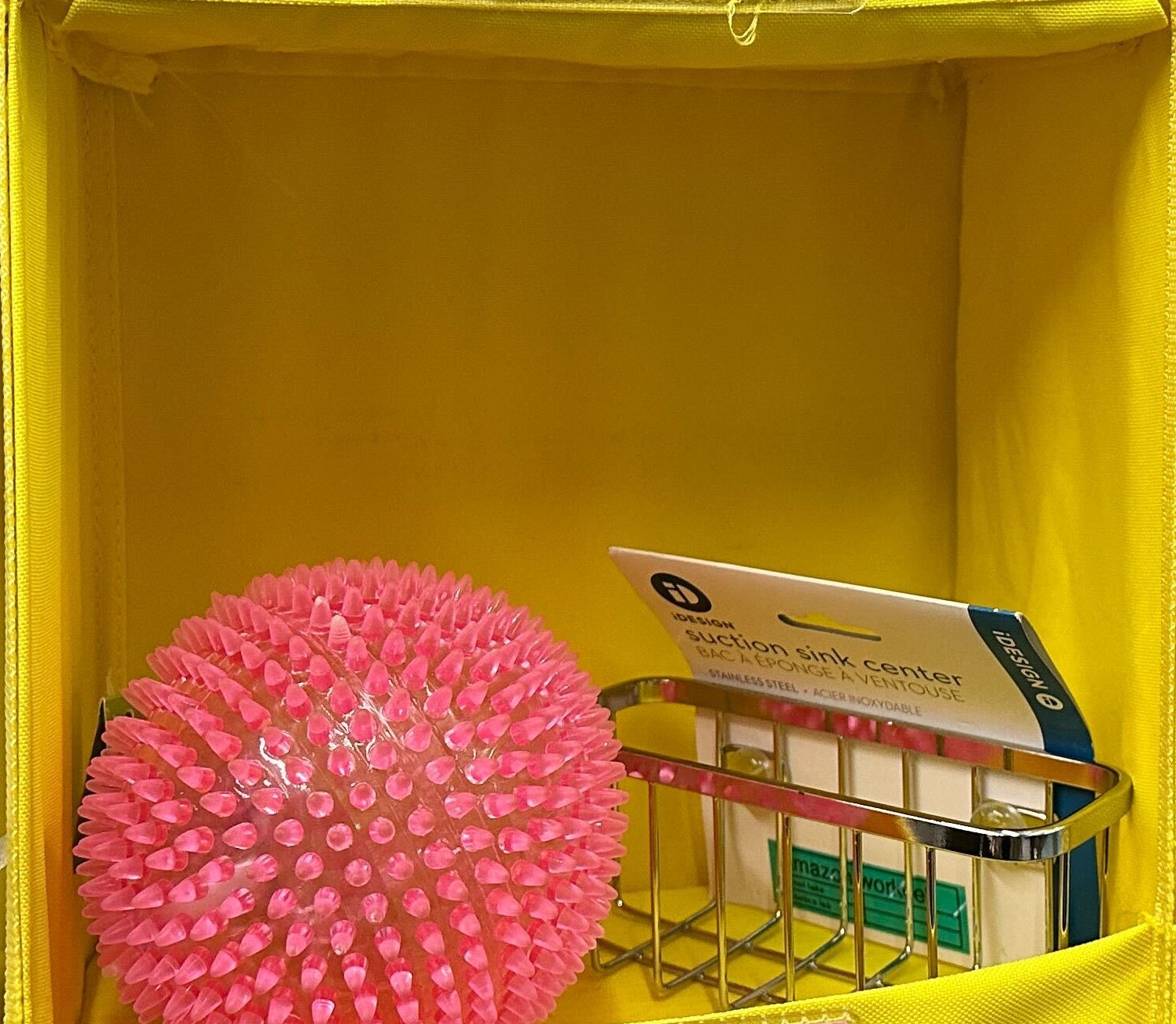}} \quad
    \subfigure[]{\centering\includegraphics[height=.12\textwidth]{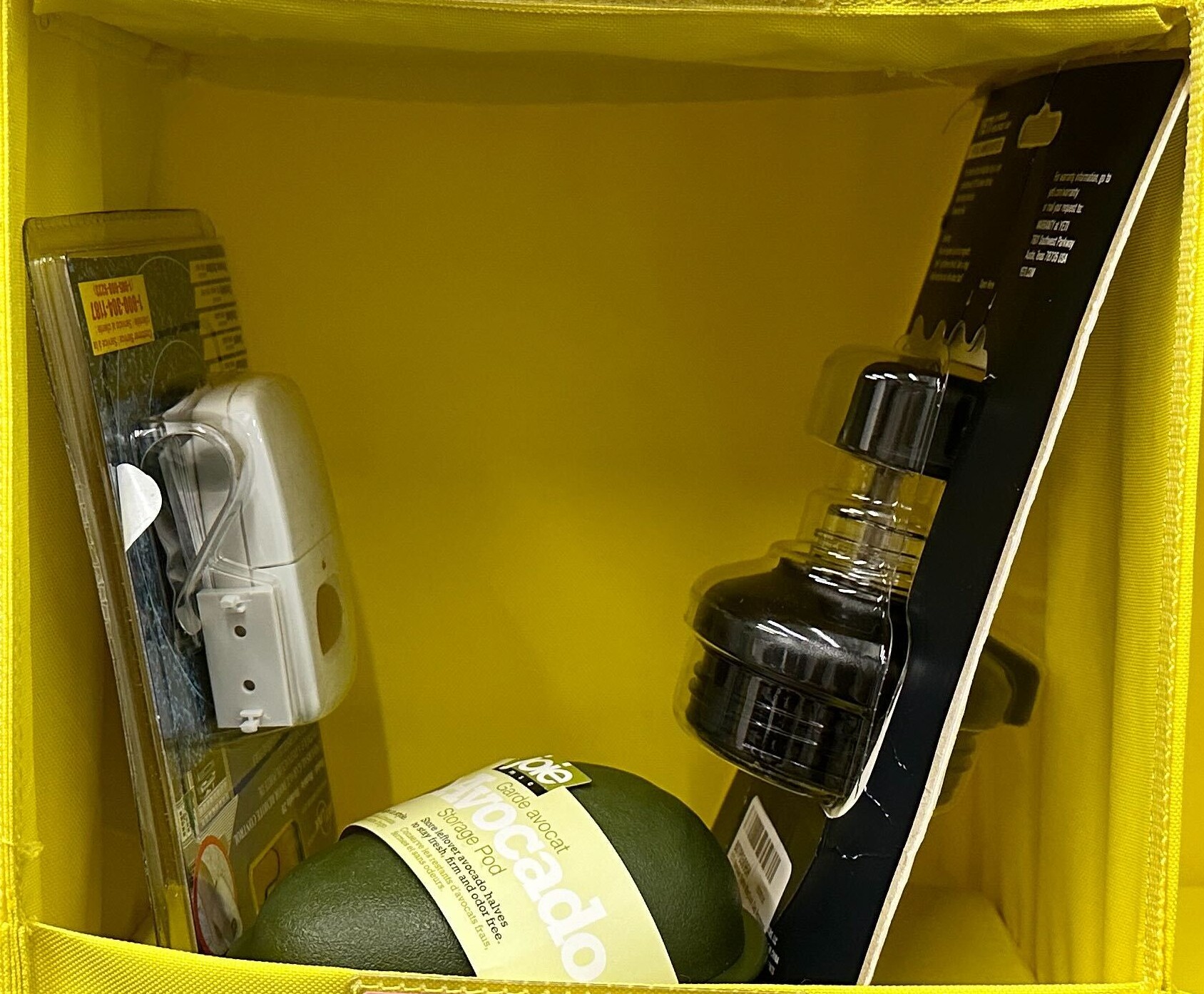}}\quad
    \subfigure[]{\centering\includegraphics[height=.12\textwidth]{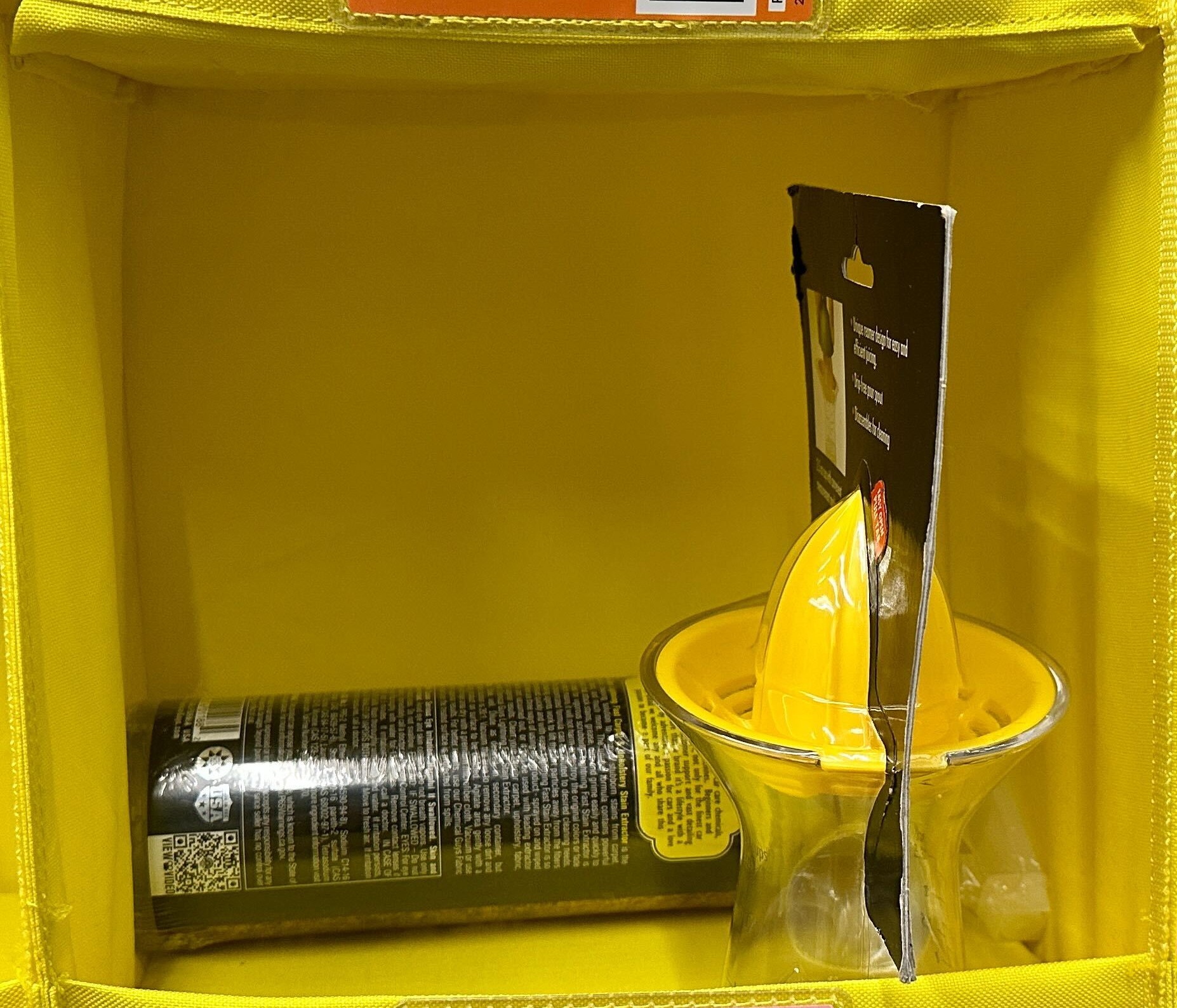}}
    \vspace{-1mm}
    \caption{\textbf{Failure cases for \optigrasp}: (a) Suction cup incompatibility with object surfaces, (b) Lack of graspable areas from a single-view, (c) Object dynamics preventing successful grasp execution}
     \label{fig:failure_cases}
    \vspace{-5mm}
\end{figure}

\autoref{fig:failure_cases} illustrates some examples of failure cases for \optigrasp. The first case (a) involves objects with porous or irregular surfaces for effective suction. The second case (b) is due to visual limitations when only a single view is available, hindering accurate identification of viable grasping points. The third case (c) involves the dynamics within the storage bin, where object movements during the grasp attempt can destabilize the grip, often preventing a secure or firm grasp and leading to failures.

In the future, we plan to explore and integrate additional foundational models and common sense reasoning from vision-language models or large language models to enhance robustness and adaptability. Furthermore, we aim to extend this method to parallel-jaw grippers and in environments to investigate the possibility of substituting depth sensors with RGB sensors in those settings.

\section{CONCLUSION}

\label{sec:conclusion}
In this study, we present a novel approach for predicting robotic suction grasping by leveraging foundation models and relying solely on RGB images, thereby bypassing the imitations of depth sensors. By harnessing pre-trained weights from the Depth Anything model and introducing the Afford Grasp head for predicting grasp affordances, our method provides an economical and effective solution for industrial warehouse picking robots. When trained solely on synthetic data, our model, \optigrasp, demonstrated robust performance in the real world and strong generalization capabilities across various objects, achieving an 82.3\% success rate through 176 successful grasps over 215 unseen objects in the real-world setup.

\vspace{-1mm}
\section*{ACKNOWLEDGEMENT}
This research is funded by the UW + Amazon Science Hub as part of the project titled “Robotic Manipulation in Densely Packed Containers.” We thank Bernie Zhu for the constructive input on paper formatting.

\bibliographystyle{IEEEtran}
\bibliography{root}

\end{document}